\pdfoutput=1

\documentclass[11pt]{article}

\usepackage{authblk}

\usepackage[]{acl}
\usepackage{color,soul}
\usepackage{times}
\usepackage{latexsym}
\usepackage{booktabs}
\usepackage{multirow}
\usepackage[T1]{fontenc}

\usepackage[utf8]{inputenc}

\usepackage{microtype}

\usepackage{graphicx}
\usepackage{amsmath}
\usepackage{booktabs} 
\usepackage{tabularx}
\usepackage{enumitem}
\usepackage{xcolor}
\usepackage{soul}
\usepackage{placeins}
\usepackage{float}
\usepackage{balance}

\usepackage{adjustbox}

\usepackage[utf8]{inputenc} 
\usepackage[T1]{fontenc}    
\usepackage{hyperref}       
\usepackage{url}            
\usepackage{booktabs}       
\usepackage{amsfonts}       
\usepackage{nicefrac}       
\usepackage{microtype}      
\usepackage{soul}

\definecolor{thistle}{RGB}{190,151,190}

\newcommand{\dataset}{\texttt{NEWTS}}

\title{NEWTS: A Corpus for News Topic-Focused Summarization\thanks{\hspace{0.1cm} The first author and the second author have an equal contribution.}}

\author[1,2]{Seyed Ali Bahrainian}
\author[1]{Sheridan Feucht}
\author[1]{Carsten Eickhoff}
\affil[ ]{\{bahrainian, sheridan\_feucht, carsten\_eickhoff\}@brown.edu}
\affil[1]{Department of Computer Science, Brown University, USA}
\affil[2]{Machine Learning and Optimization Lab, EPFL, Switzerland}

\begin{document}
\maketitle
\begin{abstract}
Text summarization models are approaching human levels of fidelity. Existing benchmarking corpora provide concordant pairs of full and abridged versions of Web, news or, professional content. To date, all summarization datasets operate under a one-size-fits-all paradigm that may not reflect the full range of organic summarization needs. Several recently proposed models (e.g., plug and play language models) have the capacity to condition the generated summaries on a desired range of themes. These capacities remain largely unused and unevaluated as there is no dedicated dataset that would support the task of topic-focused summarization.

This paper introduces the first topical summarization corpus \dataset, based on the well-known CNN/Dailymail dataset, and annotated via online crowd-sourcing. Each source article is paired with two reference summaries, each focusing on a different theme of the source document. We evaluate a representative range of existing techniques and analyze the effectiveness of different prompting methods.
\end{abstract}

\section{Introduction}\label{sec:intro}

With the recent advances in neural sequence-to-sequence models, the automatic generation of text has reached unparalleled levels of fidelity. Abstractive summarization models that aim at generating condensed versions of a source article have outperformed Lead-3 baselines on most benchmark datasets~\cite{see2017get, lewis2019bart}. However, all existing summarization benchmarks assume a one-size-fits-all paradigm under which model output is evaluated based on similarity to general-purpose reference summaries reflecting the full content of the original document. While certainly a necessary step, such evaluation approaches might not reflect the full range of summarization needs anymore. There are manifold settings in which tailored summaries matching the interests of the reader may be required. Some examples include the summarization of complex event streams with a focus on regions, entities or topics of interest for journalists or analysts, understanding reviews or opinions from different perspectives~\cite{DBLP:journals/tacl/HayashiBWANN21}, the summarization of electronic health records with a focus on the medical sub-specialty of the physician reader, or any other form of personalized summarization targeting explicitly defined or implicitly mined preference parameters. 

\begin{figure}[t]
\includegraphics[width=7cm]{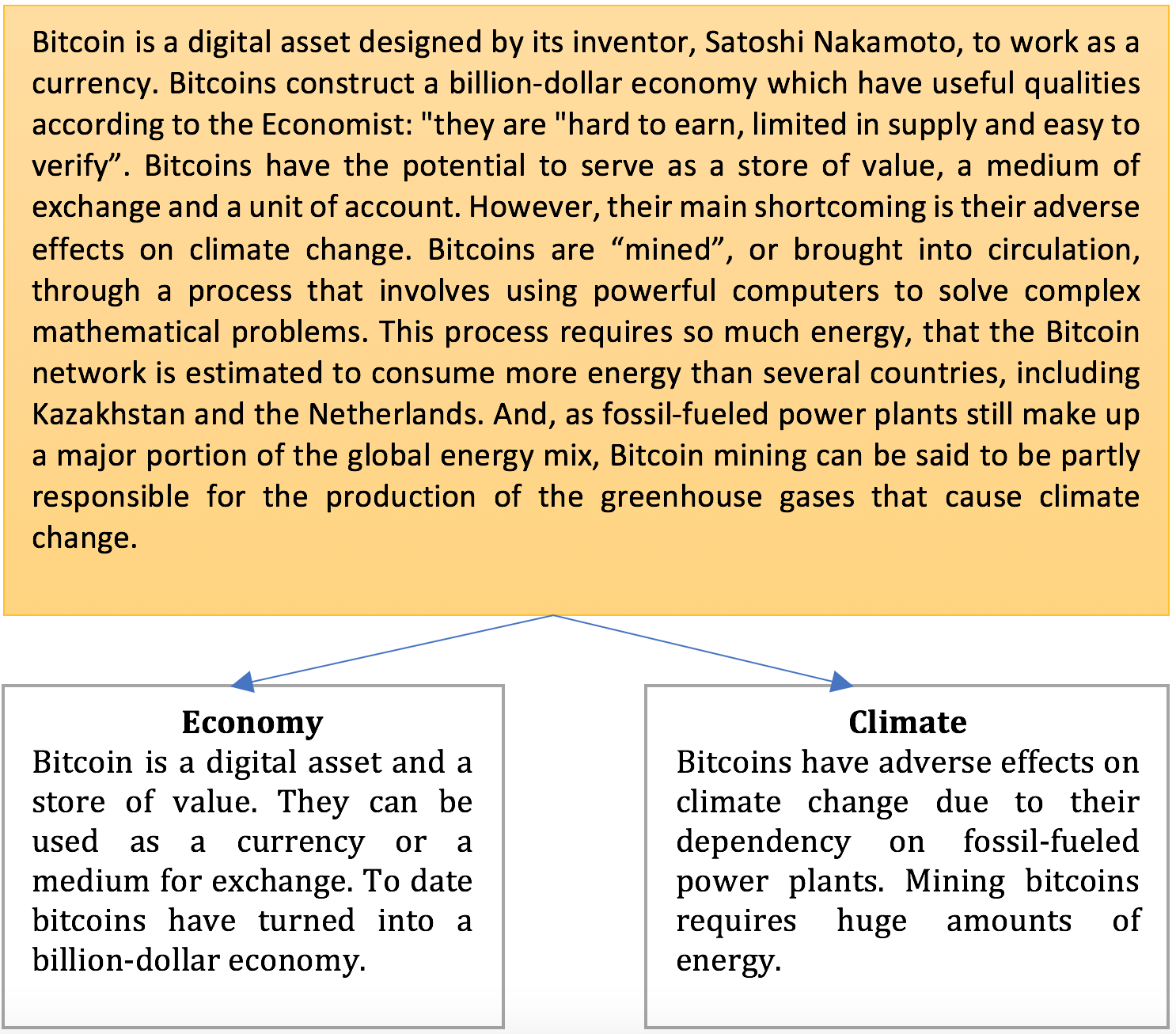}
\centering
 \caption{A topical summarization example, summarizing a sample document with respect to economy and climate topics.}
  \label{fig:example}
\end{figure}

Several recently proposed text generation models already offer the potential of steering the generation process to conform to specific topic distributions~\cite{10.1145/3464299}, or sentiment polarity~\cite{DBLP:conf/nips/ShenLBJ17}. Plug and Play Language Models (PPLM)~\cite{Dathathri2020Plug} let us condition the generation process on themes of interest and text style transfer controls selected attributes, such as politeness, emotions, or humor of the generated text~\cite{DBLP:journals/corr/abs-2011-00416}.

Despite increased efforts and interest in controlled summarization, no dataset exists on which these models can be evaluated. This paper closes this gap by introducing \dataset, a NEWs Topic-focused Summarization corpus for the controlled generation of text. It is based on documents from the well-known CNN/Dailymail dataset, to which it adds new topic-focused summaries. Figure \ref{fig:example} illustrates an article summarized with respect to two different topics. We believe that \dataset\ will significantly enrich the existing range of benchmarking collections, allowing the research community to better study and evaluate controlled text generation for summarization.

The main contributions of this paper are:

\begin{itemize}
    \item We introduce and release the first dataset of topic-based abstractive summarization\footnote{https://github.com/ali-bahrainian/NEWTS}. The dataset contains human-written topical reference summaries collected via online crowd-sourcing.
    \item We evaluate a range of existing models alongside four different prompting techniques.
\end{itemize}

The remainder of this paper is organized as follows: Section~\ref{sec:relatedwork} presents previous work on datasets for text generation. Next, Section~\ref{sec:dataset} explains the dataset collection methodology and describes the resulting corpus. Section~\ref{sec:models} discusses several existing models that we fine-tune and evaluate on the dataset. Section~\ref{sec:evaluation} presents an evaluation of these models and the various prompting strategies. Finally, Section~\ref{sec:conclusion} concludes with an outlook on future work.




\section{Related Work}\label{sec:relatedwork}






In this section, we review existing work focusing on (1) controlled text generation and (2) existing datasets in this domain. We note that this paper presents the first dataset on topic-focused abstractive summarization.

\subsection{Controlled Text Generation }

Controlled text generation encompasses transferring the style of an input text into a specific target form~\cite{DBLP:journals/corr/abs-2011-00416}. Typical Style transfer tasks in the natural language domain include shifting the formality of texts~\cite{DBLP:conf/naacl/BriakouLZT21}, the level of politeness~\cite{DBLP:conf/acl/MadaanSPPNYSBP20}, bias versus neutrality~\cite{DBLP:conf/aaai/PryzantMDKJY20}, authorship style~\cite{carlson2018evaluating}, simplicity~\cite{cao-etal-2020-expertise}, sentimental stance~\cite{DBLP:conf/nips/ShenLBJ17}, target aspects in opinion summarization~\cite{frermann-klementiev-2019-inducing, angelidis-lapata-2018-summarizing} and topical focus~\cite{10.1145/3464299}.

Persona-based text generation is another area of research that has been studied in the context of story-telling based on a particular personality type and sequences of images~\cite{chandu-etal-2019-way}.

The notion of persona-based text generation has also been studied in the context of dialogue using an Emotional Chatting Machine that generates responses in an emotional tone while conditioning on conversation history. The key feature of this work is that emotion, as opposed to persona, is deemed dynamic, and therefore emotional responses change throughout a conversation~\cite{DBLP:conf/aaai/ZhouHZZL18}.

Most of the controlled text generation tasks named above rely on learning a mapping between the source documents' latent representations and the target documents' representations. For instance, embeddings of a particular author/newspaper are learned jointly with the word embeddings of a source article and mapped onto a target form representation~\cite{fan2017controllable}.

In this paper, we focus on topic-based controlled text generation to summarize a source article around a specified topic of interest.

\subsection{Existing Datasets for Controlled Text Generation}

As explained above, datasets for different text style transfer problems exist. However, contemporary summarization models such as PPLM~\cite{Dathathri2020Plug} and CATS~\cite{10.1145/3464299} suffer from a lack of existing datasets and hence a lack of quantitative evaluation in terms of steering the topical focus in text generation. Here we review a few closely related datasets to \dataset.

The aspect-based sentiment summarization dataset WikiAsp~\cite{DBLP:journals/tacl/HayashiBWANN21} targets the generation of summaries with respect to specific points of interest. For instance, the points of interest in the case of Barack Obama (as presented in their paper) may pertain to his `early life,' career,' and `presidency.' WikiAsp is extracted automatically from Wikipedia articles, using their section headings and boundaries as a proxy for aspect annotation. Our dataset vastly differs from WikiAsp in that it covers a broader range of themes and provides dedicated human-written reference summaries while WikiAsp reverse engineers and repurposes existing articles. Finally, our dataset provides a different level of granularity and abstraction useful for separating intertwined concepts in articles. At the same time, WikiAsp merely enables the generation of text pertaining to a section header. 

Another closely related dataset is MultiOpEd, a  dataset of multi-perspective news editorials~\cite{LCUR21}. This dataset is designed around argumentation structure in news editorials, focusing on automatic perspective discovery. The assumption here is that arguments presented in an editorial typically center around a concise, focused perspective. The dataset is designed such that a system is expected to produce a single-sentence perspective statement summarizing the arguments presented. For a query on a controversial topic, two news editorials respond to the query from two opposing point-of-views constructing a lengthy statement. Each editorial comes with a single paragraph abstract plus a one-sentence perspective that abstractively summarizes the editorial's key argument in the context of the query. The query is designed to allow only two opposing arguments, i.e.~supporting or opposing it. For example, a query may be ``is it right to end the lockdown?''. Our dataset differs from MultiOpEd in that ours allows summarization of text with respect to two different (but not necessarily opposing) topics, while MultiOpEd is restricted to two opposing arguments on the same topic. 

This paper introduces and releases the first dataset on topic-focused summarization gathered via online crowd-sourcing featuring $50$ different topics. 

\section{A Novel Dataset for Controlled Summarization}\label{sec:dataset}
In this section, we present \dataset, a new dataset for controlled topic-focused text generation. We first elaborate on the steps to building the dataset. Subsequently, we present detailed statistics about the dataset. 

\begin{figure}[t]
\includegraphics[width=7cm]{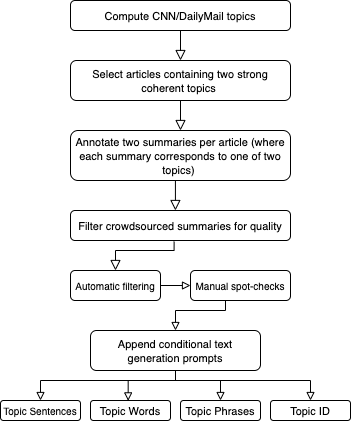}
\centering
 \caption{The step-by-step process of building the \dataset \space dataset.}
  \label{fig:process}
\end{figure}

Our dataset is built based on the well-known CNN/Dailymail dataset~\citep{hermann2015teaching, nallapati2016abstractive}, introducing an all-new facet of topical human-written summaries. For this purpose, we annotate a sample of the news articles from the CNN/Dailymail dataset via online crowd-sourcing such that each article is paired with two topic-focused human-written summaries corresponding to the top two topics present in the source article. Figure \ref{fig:process} presents the steps to creating the dataset explained in detail bellow.

\noindent\textbf{Computing Topics for the Dataset.} We begin by computing a 250-topic Latent Dirichlet Allocation (LDA)~\citep{Blei:2003:LDA:944919.944937} model on the training portion of the CNN/Dailymail dataset. LDA was selected due to convenience of use, and $k=250$ topics empirically showed best coherence and consistency among various choices in the $k \in [50,300]$ range. From this model, we manually discard noisy or uninformative topics, keeping only the top 20\% (50 topics) with the highest Normalized Point-wise Mutual Information (NPMI) coherence score~\citep{bouma2009normalized}. We perform this aggressive pruning of topics out of feasibility considerations regarding the number of documents per topic provided for fine-tuning neural summarization models. A list of all $50$ topics is presented in the appendix. 

\noindent\textbf{Selecting articles for annotation.} After computing the $50$ target topics of the dataset, we search the CNN/Dailymail dataset for source articles containing at least two topics from the pool of $50$ topics with a topic prevalence above an empirically determined threshold. 


By identifying documents that contain at least two topics with a topic prevalence above the empirical threshold $0.1$, and a cumulative probability of both topics above $0.30$, we ensure that the main content of the source article can be captured by focusing on the two main topics. Consequently, each source article will be summarized twice, with each summary concentrating on one of the main two topics.

\begin{table*}[ht]
\small
\vspace{6pt}
\begin{tabularx}{\linewidth}{XXXX}

\multicolumn{1}{c}{\textit{\textbf{Topic Words}}} &  \multicolumn{1}{c}{\textit{\textbf{Topic Phrases}}} & \multicolumn{1}{c}{\textit{\textbf{Topic Sentence}}}  & \multicolumn{1}{c}{\textit{\textbf{Topic ID}}}   

\tabularnewline  \bottomrule

 court, judge, case, appeal, justice, order, ruling, ruled, magistrates, ordered & a court ruling, department of justice, appealed against a court ruling, judge reviewing a case, court order, magistrates & This topic is about a court ruling, department of justice, appealing against a court ruling, judge reviewing a case, a court order, and magistrates. & \_TID78 
 \tabularnewline \bottomrule
\hline
fire, residents, san, wood, firefighters, burning, burned, blaze, flames, fires & firefighters tackled the blaze, wood burning, residents evacuating, flames, spit embers downwind, burning buildings      & This topic is about firefighters tackling the blaze, wood burning, residents evacuating, flames, spit embers downwind, and burning buildings.        & \_TID153
 \tabularnewline \bottomrule
 
\end{tabularx}
\caption{Two topic examples with their corresponding topic phrases, topic sentences, and topic IDs }\label{prompts}

\end{table*}

\begin{figure}[!tbp]
  \centering
  \begin{minipage}[b]{0.52\textwidth}
    \includegraphics[width=\textwidth]{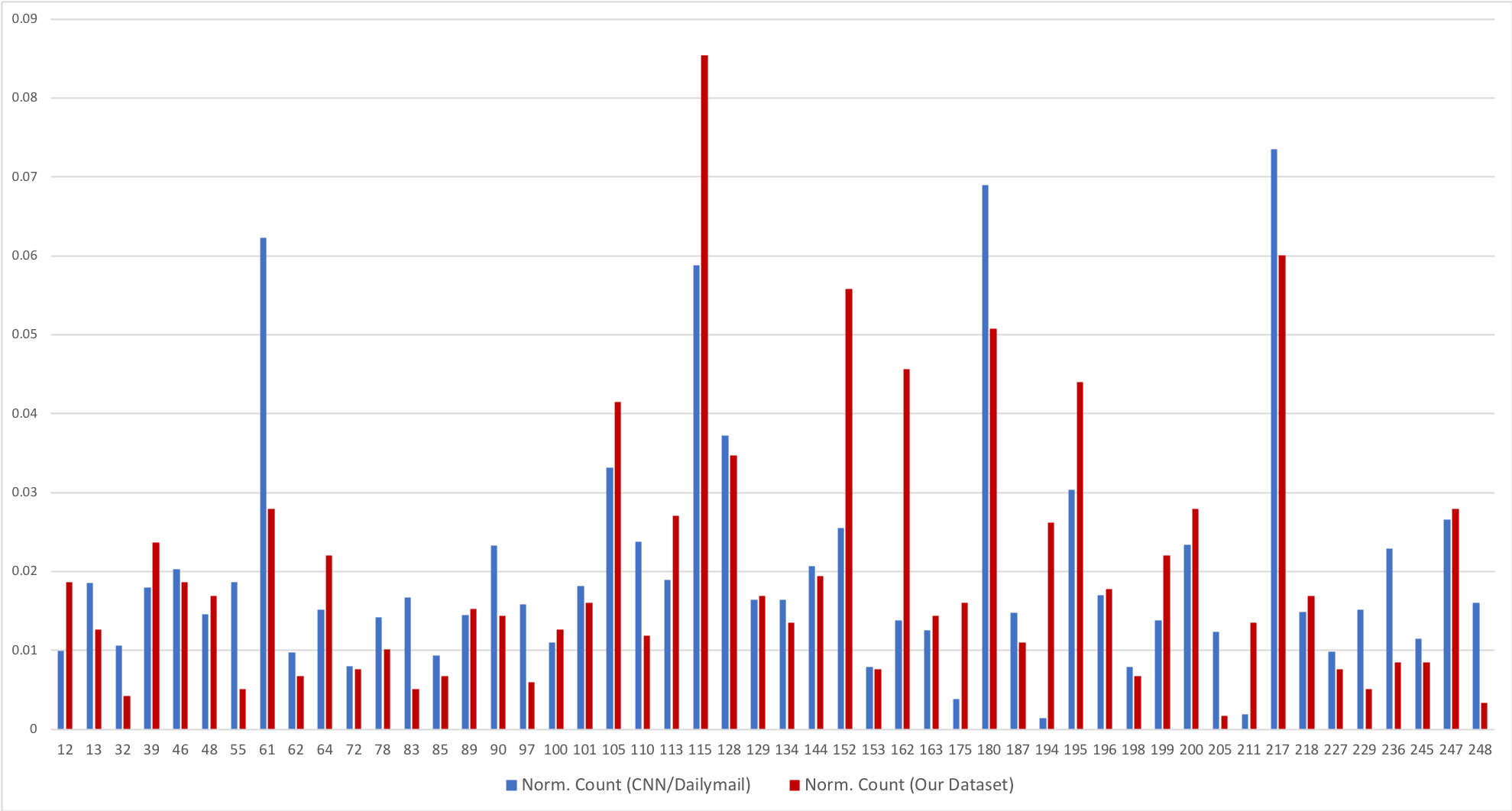}
    \caption{Comparison of per topic normalized counts of \dataset\ test documents versus CNN/Dailymail  counts}\label{stats1}
  \end{minipage}
  \hfill
  \begin{minipage}[b]{0.52\textwidth}
    \includegraphics[width=\textwidth]{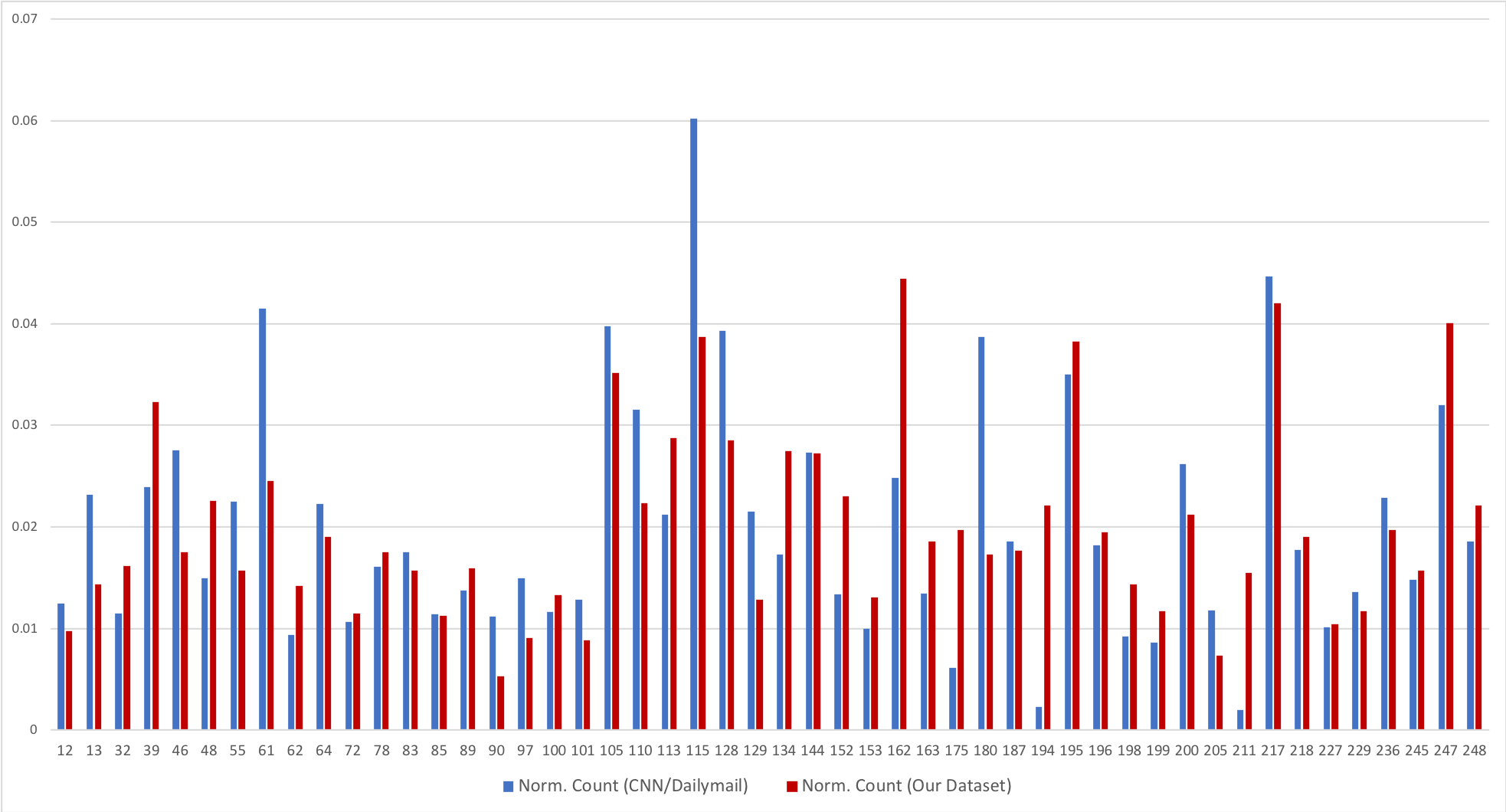}
    \caption{Comparison of per topic normalized counts of Train Documents of our Dataset versus CNN/Dailymail}\label{stats2}
  \end{minipage}
\end{figure}

\noindent\textbf{Annotating each source article with two topic-focused summaries.} We use Amazon MTurk to obtain two summaries of the same source article, each focused on a different topic. The annotation process is designed such that a crowd-sourcing worker receives a source article and two topics written in the form of hand-curated phrases, along with instructions on how to write two summaries about the source article. The instructions request having at least three sentences per summary, focusing on one topic while avoiding the other topic as much as possible without any copy-pasting of entire sentences. For each of the $50$ most coherent topics used in the dataset, we display the top $20$ words with the highest probability of being present in that topic and manually write a series of phrases separated by commas exemplifying the topic in a few words. 

\noindent\textbf{Controlling the quality of the human-written summaries.} Once the human-written summaries are obtained, we perform a quality check on them to reject noisy annotations from the dataset. To ensure the dataset's quality, (1) we use a validated script to filter out unacceptable summaries automatically and (2) perform manual spot checks and ban problematic workers to further reduce potential noise in the dataset. We explain each of these steps below:

The automatic filtering script is developed to identify and reject summaries that are too short (i.e., shorter than three sentences required from the workers) or do not form a grammatical sentence, summaries that are not related to the topics discussed in the source article, summaries that do not mention the same entities discussed in the source article (using named entity recognition) and summaries that contain exact copy-pasting of full sentences from the source article. To check the topics of the summary and compare them with that of the source article, the script uses the same LDA topic model described earlier in this section. Subsequently, the script is validated by conducting three pilot studies, each annotating $100$ documents, bringing the total number of documents tested to $300$. We manually assess each annotation in order to evaluate the script. In the third pilot study, our script reached $100 \%$ agreement with two independent human experts in terms of accepting/rejecting the annotations. 

We still conduct manual spot checks of the script output throughout the crowd-sourcing process to ensure a high-quality dataset. One of the two human experts read each sampled annotation and determine whether the quality satisfies the task description and the criteria explained earlier and rejects those annotations that do not meet the requirements. We use a z-test with a $95\%$ confidence level and an error margin of $+/- 9.24\%$ (i.e.,  from $85.76\%$ to $100\%$ of our population) as our sampling technique. Therefore, with a confidence of $95\%$, high quality for the annotations is ensured.

\noindent\textbf{Designing prompts for conditional text generation. }In order to be able to condition a generation process sequence-to-sequence models on certain topics for producing summaries, we design four different prompt types paired with each summary to allow advanced prompt engineering techniques. In the following, we explain each method:

\begin{enumerate}
    \item Topic Words: the first prompting technique utilizes the top $10$ words based on their probability assignment in that topic separated by commas.
    
    \item Topic Phrases: the second prompting method consists of the exact topic phrases that were hand-written based on the top topic words and sent to the annotators to understand the topic.
    
    \item Topic Sentences: the third prompting method is a hand-written sentence describing a topic and what that topic is about. In practice, such sentences connect all the topical phrases from the previous prompting method in a sentence.
    
    \item Topic ID: the fourth prompting method represents each topic with a unique topic identifier to examine the possibility of learning a topic embedding using a simple topic identifier. 
\end{enumerate}

Table \ref{prompts} presents two of the $50$ sample topics in the first column with their top $10$ corresponding words according to their associated probability in that topic. The first topic is related to \textit{courts and justice} while the second topic is related to \textit{fires and burning residences}. The four columns of the table correspond to each prompt type described above.

Each of the prompts presented in the paper are prepended to the tokens of the source article separated by a special separation token and fed to the Transformer-based models. We will compare all these prompting methods in a benchmark for the task of topic-controlled abstractive summarization.

The resulting dataset consists of 3,000 source articles (2,400 from the training set of the CNN/Dailymail dataset to construct the train set of \dataset, and $600$ articles from the test set of the CNN/Dailymail dataset to form the test set of \dataset). Each article is annotated with two summaries, each focusing on a different topic present in the article. The overall number of manually composed topical summaries is, therefore, $6,000$ ($4,800$ for training and $1,200$ for testing). The summaries of the final training set have a length of $416.1$ characters on average, while the average number of sentences and number of tokens per summary is 5.5 and 70.2, respectively. The average number of characters per test summary is $412.9$, while the average number of sentences and the average number of tokens per summary are $5.0$ and $70.1$, respectively. 

Figures \ref{stats1} and \ref{stats2} show the number of documents per topic normalized by size present in our dataset side-by-side that of the CNN/Dailymail dataset. The former figure illustrates these numbers for the test sets, while the latter pertains to the train sets. 

\section{Topical Summarization Models}\label{sec:models}

\noindent\textbf{Text-to-Text Transfer Transformer: }The T5 (Text-to-Text Transfer Transformer) model is an important example of the Transformer family~\citep{raffel2019exploring} that uses transfer-learning on the original Transformer architecture~\cite{vaswani2017attention}. The authors study several variants of the Transformer architecture and finally fine-tune them on different natural language processing tasks. The main difference from the original model is the use of relative positional embeddings as an explicit position signal of the tokens.

\noindent\textbf{BART: }The next model that is noteworthy in this domain is BART~\cite{lewis2019bart}. BART is a denoising autoencoder for pretraining sequence-to-sequence natural language processing models. It is trained by ``corrupting text with an arbitrary noising function and learning a model to reconstruct the original text''~\cite{lewis2019bart}. Analogous to the T5 model, BART is based on the Transformer architecture~\cite{vaswani2017attention}. It uses a number of noising approaches, such as token masking, token deletion, randomly shuffling the order of the original sentences, and a novel in-filling scheme, where spans of text are replaced with a single mask token. The only major difference to the Transformer architecture is that, following GPT, the authors replace ReLU activation functions with GeLUs~\cite{hendrycks2016gaussian}. They also state that their proposed architecture ``is closely related to that used in BERT, with the following differences: (1) each layer of the decoder additionally performs cross-attention over the final hidden layer of the encoder (as in the transformer sequence-to-sequence model); and (2) BERT uses an additional feed-forward network before word prediction, which BART does not''~\cite{lewis2019bart}. BART is then fine-tuned on in-domain data for text generation tasks such as abstractive summarization.

\noindent\textbf{ProphetNet: }The final model in this category is ProphetNet~\cite{yan2020prophetnet}, which currently represents the state-of-the-art in abstractive summarization. This model also utilizes the Transformer architecture~\cite{vaswani2017attention}. The main feature of ProphetNet is changing the original sequence-to-sequence optimization problem of predicting the next single token into predicting the $n$ next tokens simultaneously. The authors show that this approach outperforms all other baselines in abstractive summarization in terms of ROUGE scores. 

\noindent\textbf{Plug and Play Language Models: } The Plug and Play Language Model (PPLM)~\cite{Dathathri2020Plug} is based on GPT-2 using the same original Transformer architecture~\cite{vaswani2017attention} as the models above. PPLM uses GPT-2 for text generation. However, it comes with an attribute model that conditions the generation process on given or previously generated text. The attribute model is fed with a bag of words signaling the target topical focus to the model. 

\noindent\textbf{Customizable Abstractive Topic-based Summarization: }Finally, we include the Customizable Abstractive Topic-based Summarization (CATS)~\citep{10.1145/3464299} model as an example of pre-Transformer seq-to-seq models based on LSTMs. The encoder-decoder architecture has Bidirectional LSTMs as the encoder and an LSTM network as the decoder. The model utilizes attention weights governed by an LDA topic model to modify the attention weights of the input tokens as represented by the encoder based on their topic assignment. This process utilizes a set of pre-defined topics derived from target summaries to learn the topics the output text should cover. 

\section{Evaluation}\label{sec:evaluation}

\noindent\textbf{ROUGE Evaluation of all Models.} In the first experiment we evaluate the various models on our new dataset in terms of  $F_{1}$ ROUGE 1, $F_{1}$\ ROUGE 2, and $F_{1}$ ROUGE L scores using the official Perl-based implementation of ROUGE~\cite{lin2004rouge}.

Table \ref{benchmark} presents the results of this experiment. We compute the optimal number of epochs and the beam size for decoding via 3-fold cross-validation for each model. In the table, `b' after a model name indicates a `base' model size while `L' indicates a `large' model size. Additionally, `T-W' indicates the prompt `topic-words,' `T-ph' indicates a `topic-phrase' prompt, `T-Sent' indicates a `topic-sentence' prompt, `no prompt' means no prompting was used while fine-tuning a model, and `CNN-DM' indicates that the model was fine-tuned on the same source articles of our dataset paired with their original corresponding CNN/Dailymail summaries. The initial goal of this experiment is to probe whether the model variations with any of the topical prompts can outperform the `no prompt' versions, which are trained on \dataset\ without conditioning on a topical prompt and the `CNN-DM' versions, which are trained for a standard summarization task. 

As we observe in the table, in the case of `BART-b,' `T5-b', `T5-L' as well as `ProphetNet,' the model variations with topical prompts outperform both the `no prompt' version as well as the `CNN-DM' version in terms of the ROUGE scores. We do not observe a conclusive pattern when comparing the different prompting methods in terms of the ROUGE scores. That is, there is no one prompt that leads to a higher ROUGE performance for all models. 

\begin{table}[h]
\centering
\small
\begin{tabular}{c|c|c|c|c}
                      & R1 & R2 & RL & Topic Focus \\ \hline
BART-b + T-W      & 31.14  &  10.46 &  19.94 & 0.1375\\ 
BART-b + T-Ph     &  31.01  &  10.36 & 19.91  & 0.1454\\ 
BART-b + T-Sent  & 30.38  &  09.70 & 19.48  &  0.1513\\ 
BART-b T-ID         &  30.97 & 10.23  &  20.08 &0.1399 \\ 
BART-b no prompt    & 16.48  &  0.75 & 11.71  & 0.0080\\ 
BART-b CNN-DM    & 26.23  &  7.24 & 17.12  & 0.1338 \\ 
T5-b + T-W        &  31.78 & 10.83  &  20.54 & 0.1386\\ 
T5-b + T-Ph      &  31.55 &  10.75 &  20.27 & 0.1426\\ 
T5-b + T-Sent    &  31.40 &  10.37 & 20.35  & 0.1528\\ 
T5-b + T-ID           &  31.44 & 10.64  & 20.06  & 0.1342\\ 
T5-b no prompt         & 30.98  & 10.19  &  20.23 & 0.1379 \\ 
T5-b CNN-DM         & 27.87  & 8.55  & 18.41  & 0.1305\\ 
T5-L + T-W        & 30.92  & 10.01  &  20.19 & 0.1598\\ 
T5-L + T-Ph      &  31.40 &  10.50 &  20.27 &  0.1457\\ 
T5-L + T-Sent    &  30.64 & 09.84  & 19.91  & 0.1462\\ 
T5-L + T-ID           & 30.35  &  9.93 &  19.77 & 0.1335\\ 
T5-L no prompt         &  30.06 & 9.55  & 19.25  & 0.1366 \\ 
T5-L CNN-DM         & 28.44  & 8.49  &  18.61 & 0.1286 \\ 

ProphetNet + T-W        &  31.91 &  10.80  & 20.66  &  0.1362\\ 
ProphetNet + T-Ph      &  31.56 & 10.35 & 20.17  & 0.1474 \\ 
ProphetNet + T-Sent    & 31.40  & 10.03  &  20.02 & 0.1633 \\ 
ProphetNet no prompt         &  30.22 & 9.67 & 19.27  & 0.1316 \\ 
ProphetNet  CNN-DM         &  28.71 & 8.53  & 18.69  & 0.1295 \\  

PPLM         & 29.63  &  9.08 &  18.76 & 0.1482\\ 
CATS         & 30.12  & 9.35  & 19.11  & 0.1519 \\ 
\end{tabular}
\newline
\caption{Benchmark comparing various models and prompting methods, using a 3-fold cross validation in terms of $F_{1}$ ROUGE 1, $F_{1}$\ ROUGE 2, and $F_{1}$ ROUGE L and the LDA topic-focus score.}\label{benchmark}
\end{table}

As a result, we conclude that while the topical prompts do lead to performance improvement on the topic-focused summarization task, we do not observe a conclusive superiority pattern among the prompts in terms of the ROUGE performance.

\noindent\textbf{Evaluating the Topicality of Output Summaries.} In the second experiment, we evaluate the topical focus of the generated summaries by each model in terms of the topic probability score computed by the LDA topic model, indicating the strength of a target topic presence. Therefore, we design an experiment to assess the performance of the different models with different prompt types in how topic-focused their output summaries are. For this purpose, we utilize the LDA topic model to compute a per target-topic score in each generated summary. Then we compute the average of this score across all generated summaries for their corresponding pre-defined target topic. We expect the models using topical information to have a higher topic\_focus score. We present the results of this experiment in the right-most column of Table \ref{benchmark}. From the results of this experiment, we observe that in all cases, the topical prompt variations of each model outperform the `CNN-DM' variation indicating that the models trained for topic-focused summarization produce summaries that are more target-topic-oriented.

Subsequently, we observe that topic sentence prompts outperform all other prompting techniques in achieving a high LDA target-topic score, suggesting that topic sentence prompting provides models with superior topic context information.

\begin{figure}
  \centering
  \includegraphics[width=0.95\linewidth]{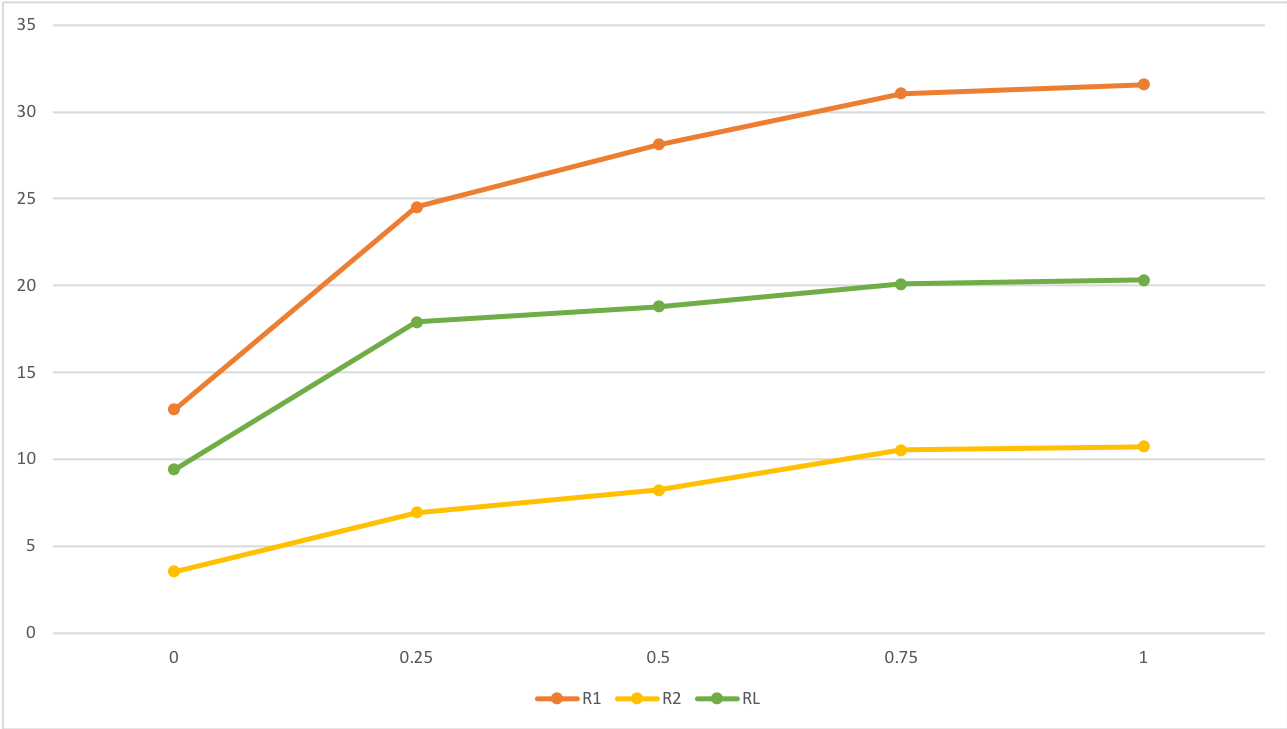}
  \caption{Figure showing the impact of training data size on ROUGE performance comparing performance of T5-base + Topic\_Phrases fine-tuned with $25\%$, $50\%$, $75\%$ and $100\%$ of the training data}
  \label{fig:size}
\end{figure}

\noindent\textbf{Evaluating the Effect of Training Data Size on Performance.} In this experiment, we investigate the effect of training data size on ROUGE performance. For this purpose, we experiment with the T5-base model and fine-tune it first on $25\%$ of the training data, then on $50\%$, on $75\%$, and finally on all the data to analyze the effect of training data size on ROUGE scores. Figure \ref{fig:size} illustrates the results of this experiment. The figure shows that increasing the training set size from $25\%$ to $75\%$ results in a significant improvement in performance in terms of ROUGE while increasing the dataset size from $75\%$ to $100\%$ indicates a convergence. The findings in this experiment indicate that the model improves in ROUGE performance scores as we increase the training data size up to $75\%$ showing a desirable behavior. Moreover, the performance curves converge after $75\%$, implying a sufficient dataset size.

\noindent\textbf{A Qualitative Human Study of Topicality on the Dataset.} This experiment assesses the dataset quality in terms of the topical focus of the summaries. To achieve this, we design a survey with three human judges. We randomly select $100$ articles from our dataset to conduct the user study. Subsequently, for each article, we present one of its topical summaries, the target topic of the summary, and the standard non-topical summary of the article from the original CNN/Dailymail dataset. The human judges are asked to identify the topical summary among the two options given the target topic. Therefore, the judges can make a binary decision determining the topic-focused summary. The results of this experiment reflect that with an accuracy of $93\%$, the judges identify the topical summary. The Kappa agreement score between the three judges was $0.7845$. The findings of this experiment suggest that the quality of the dataset in terms of the summaries' topical focus is very high.

\begin{figure}
  \centering
  \includegraphics[width=0.98\linewidth]{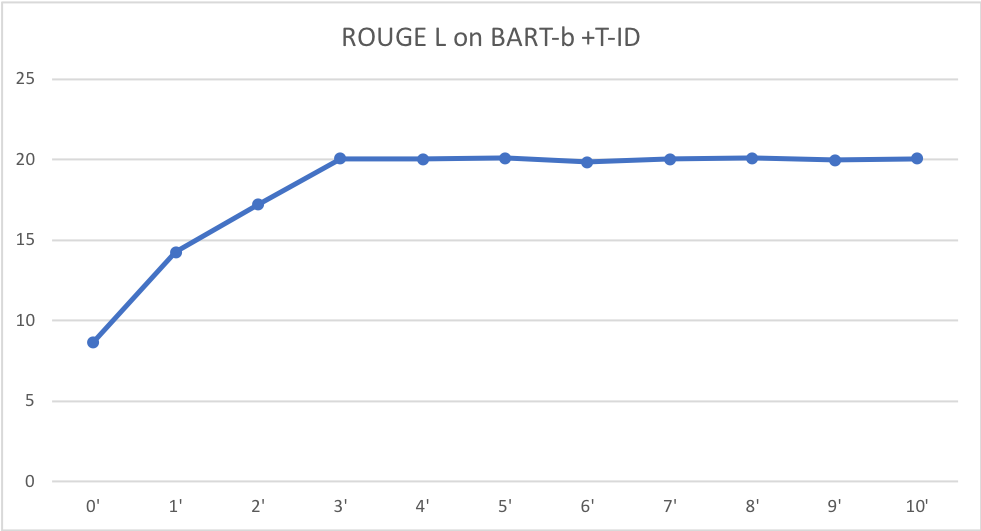}
  \caption{Figure showing the impact of the fine-tuning epochs of the BART-b + T-ID model on ROUGE L performance.}
  \label{fig:epoch}
\end{figure}

\begin{table}[h]
\begin{center}
\tiny
\begin{tabular}{ | m{2cm} | m{20em} | } 
  \hline
  Source of Summary & Summary Text \\ 
  \hline
  Ground Truth Summary1 (Pop Music)&  After experiencing some terrible customer service on an airline, a band wrote a sarcastic song about the experience. It became a hit, notably among other passengers on that airline. However, not everyone is impressed with their musical talent and lyrics.  \\ 
  \hline
  Ground Truth Summary2 (Airline) & Ryanair is well-known for upsetting its passengers. Its flight attendants are known to be rude and its surcharges are ill-received. It is launching a marketing campaign to revamp its image. \\ 
  \hline
  
    BART-b + Topic-Sent (\textcolor{red}{Pop Music}) & \textcolor{red}{Sidonie}, a well-known band from Catalonia created a tongue-in-cheek \textcolor{red}{song} during a Ryanair flight to Santiago de Compostela in northern Spain. The \textcolor{red}{band's} lead \textcolor{red}{singer strummed} a \textcolor{red}{ukulele} as the group \textcolor{red}{serenades} passengers on board - despite Ryanair's scathing review of the sarcastic \textcolor{red}{lyric}. The \textcolor{red}{lyrics} read: 'Ryanair, how we like Ryanair, and its flight attendants, they could not be nice,' from Ryanair. \\ 

\hline  
    BART-b + T-W (\textcolor{blue}{Airline}) & A well-known band, created a somewhat ironic song during a recent Ryanair \textcolor{blue}{flight} to Santiago de Compostela in northern Spain.
  Explaining the motivation behind their hilarious video, the band wrote online: 'After repeatedly receiving the usual lack of respect shown by \textcolor{blue}{flight attendants} on the Ryanair route to Santiago, we were moved to compose and perform a song dedicated to them. Other \textcolor{blue}{passengers} also complained about the \textcolor{blue}{airline}. \\

  \hline
  \hline
    Ground Truth Summary1 (Presidential Election) & Marco Rubio is running for president. The Florida Senator is already receiving large contributions for his campaign from donors. He will need the money, as he is also competing with Republican candidates who also have received large donations.\\
  \hline
   Ground Truth Summary2 (Marriage and Civil Law) & Marco Rubio claims that people are born gay or straight, rather than being influenced by outside circumstances. He supports people's right to choose, even though he himself does not agree with gay marriage. He does say that the legality of gay marriage should be decided by state legislators rather than the court system.
 \\ 
  \hline
  
  T5-L + T-Sent (\textcolor{green}{Presidential Election}) & Senator Marco Rubio announced he is running for \textcolor{green}{president} last week. Donors have said their \textcolor{green}{candidate} has already received monetary commitments in excess of the \$40 million he will likely need to battle through a \textcolor{green}{presidential primary} season that will feature a crowd of seasoned \textcolor{green}{Republican candidates} with strong financial backing. \\

  \hline

 T5 + T-ph (\textcolor{pink}{Marriage and Civil Law}) &  Marco Rubio believes that people are born with a sexual preference while insisting state \textcolor{pink}{legislators} should decide whether or not to allow gay \textcolor{pink}{marriage}. The presidential candidate spoke to CBS' Face the Nation after admitting in an interview he would attend the same-sex \textcolor{pink}{wedding} of a family member or staffer - even if he didn't agree with the decision. The Florida Senator told Bob Schieffer that he wasn't against gay \textcolor{pink}{marriage}, but believes the 'definition of the institution of \textcolor{pink}{marriage} should be between one man and one woman'. \\
  \hline
\end{tabular}\caption{Two sets of qualitative examples of Ground-truth summaries alongside system-generated summaries. Change of a target topic results in a significant vocabulary shift shown in color.}\label{tab:qualitative}
\end{center}
\end{table}

\noindent\textbf{Analyzing the Number of Fine-tuning Epochs on ROUGE Performance.} In this experiment, we test the learnability of the abstractive topic-focused summarization task by a Transformer model. To achieve this, we examine the effect of the number of fine-tuning epochs on performance gain. For this purpose, we randomly select one of the model variations presented in Table \ref{benchmark}, namely `bart-b + T-ID,' and analyze it in terms of its learning behavior in terms of the ROUGE L performance metric over different epochs. The results of this experiment shown in Figure \ref{fig:epoch} suggest that through the first three epochs, the model learns the topic-focused summarization task and finally converges with minimal performance differences on the higher number of epochs. We conclude that in three epochs, the `bart-b + T-ID' model learns topic-focused summarization and shows a convergence behavior.

\noindent\textbf{Qualitative Examples from the dataset and Model Outputs.} Finally, we present randomly selected qualitative examples from the dataset along with the outputs generated by different models showing the quality of topic-conditioned text generation. The sample outputs presented in Table \ref{tab:qualitative} demonstrate high quality in summarizing an article with respect to two different topics.

\section{Conclusions and Future Work}\label{sec:conclusion}

This paper designs and releases the first publicly available dataset for controlled topic-focused abstractive summarization, \dataset. Our dataset encompasses four prompt types to allow various conditional text generation techniques.

We showed through extensive experimentation that the new dataset is of high quality. We believe that this dataset will serve the community to advance research in controlled text generation and topical summarization as a foundation for future research.

Our findings indicate that the sequence-to-sequence Transformer baselines fine-tuned with topical prompting outperform the non-topical variation model counterparts showing that the models do learn topical representations for a topic-focused text generation. Additionally, our experiments suggest that topical sentence prompts surpass other prompt types in steering the generation process to achieve a high LDA target topic score. This finding is in line with the notion that contextual language models learn better sentence representations than other word constructions, such as the other different prompt types proposed in this paper.

In the future, we plan to design a topic-focused generative model that not only would condition the generation process on a pre-defined topic but would also penalize the generation of non-target-topic words in the decoding phase. Furthermore, we plan to investigate the problem of live topic-focused text generation in a zero or few-shot learning process using the new \dataset \space dataset.

\section*{Acknowledgements}
This research is supported in part by the NSF (IIS-1956221), SNSF (P2TIP2\_187932), ODNI and IARPA via the BETTER program (2019-19051600004). The views and conclusions contained herein are those of the authors and should not be interpreted as necessarily representing the official policies, either expressed or implied, of NSF, SNSF, ODNI, IARPA or the U.S.\ Government.

\bibliography{anthology,custom}
\bibliographystyle{acl_natbib}
\clearpage

\appendix

\section{Appendix: NEWTS Topics}
\label{sec:appendix}

The following table presents all 50 topics covered in the NEWTS dataset using the top five words present in each LDA topic. As it can be seen, our newly introduced dataset encompasses a vast range of coherent topics present in the real-world news domain. We have presented each topic with its original topic id as obtained from the LDA model to facilitate the reproducibility of the results presented in this paper. Furthermore, we plan to release the dataset and our entire code base to ensure the reproducibility of our experiments.
\begin{table}[]
\small
\begin{tabular}{r|l}
Topic Id & Topic Words                                         \\ \hline
62                            & island, beach, sea, gaal, navy                       \\
32                            & water, river, lake, bridge, walker                   \\
78                            & court, judge, case, appeal, justice                  \\
46                            & law, legal, state, marriage, rights                  \\
12                            & islamic, terror, terrorist, al, threat               \\
229                           & hotel, guests, bar, glass, wine                      \\
105                           & charged, allegedly, charges, arrested, alleged       \\
72                            & health, virus, cases, people, bird                   \\
153                           & fire, residents, san, wood, firefighters             \\
97                            & visit, pope, peace, catholic, roman                  \\
134                           & air, plane, aircraft, flight, flying                 \\
13                            & price, cost, products, market, prices                \\
187                           & website, disease, spread, ill, contact               \\
152                           & united, manchester, liverpool, chelsea, league       \\
195                           & court, trial, guilty, prison, heard                  \\
64                            & group, forces, fighters, killed, fighting            \\
113                           & campaign, clinton, governor, presidential\\
163                           & airport, passengers, flight, travel, airlines        \\
162                           & president, obama, white, house, barack               \\
199                           & cup, real, madrid, brazil, ronaldo                   \\
129                           & attack, attacks, killed, attacked, bomb              \\
175                           & house, committee, congress, senate, republican       \\
211                           & london, british, uk, britain, royal                  \\
227                           & music, singer, song, band, bruce                     \\
194                           & russian, russia, european, europe, ukraine           \\
217                           & club, team, season, players, england                 \\
61                            & match, murray, won, title, round                     \\
90                            & arsenal, ball, alex, wenger, villa                   \\
115                           & family, wife, daughter, husband, couple              \\
236                           & film, movie, character, films, viewers               \\
89                            & weight, pounds, fat, diet, body                      \\
39                            & war, military, defence, army, iraq                   \\
180                           & goal, win, side, scored, minutes                     \\
247                           & tax, average, benefits, people, rate                 \\
110                           & billion, figures, economy, global, growth            \\
85                            & coast, miles, storm, east, map                       \\
196                           & school, schools, teacher, high, education            \\
248                           & hospital, medical, doctors, patients, care           \\
205                           & art, museum, display, century, history,              \\
83                            & road, driver, driving, traffic, speed                \\
48                            & food, restaurant, eat, eating, babies                \\
144                           & online, users, internet, site, device                \\
100                           & earth, sun, climate, planet, change                  \\
200                           & children, child, parents, birth, born                \\
198                           & study, researchers, google, scientists, university   \\
245                           & facebook, mobile, phone, network, samsung            \\
128                           & money, pay, paid, card, credit                       \\
55                            & energy, power, heat, plant, fuel                     \\
101                           & crown, grand, race, hamilton, team                   \\
218                           & snow, weather, cold, winter, temperatures           
\end{tabular}
\caption{First five words (i.e. assigned the highest probability in the LDA topic) for each of the entire 50 topics covered in \dataset \ dataset.}
\label{tab:my-table}
\end{table}
\end{document}